\documentclass[10pt,twocolumn,letterpaper]{article}
\usepackage{cvpr}
\usepackage{times}
\usepackage{epsfig}
\usepackage{graphicx}
\usepackage{amsmath}
\usepackage{amssymb}
\usepackage{cite}
\usepackage{makecell}
\usepackage{algorithm} 
\usepackage{algpseudocode} 
\usepackage{listings} 
\usepackage{xcolor} 

\begin{document}

\title{La-SoftMoE CLIP for Unified Physical-Digital Face Attack Detection}
\author{
	Hang Zou$^{\rm 1}$\footnotemark[1], 
	Chenxi Du$^{\rm 2, \rm 3}$\thanks{Equal contributions.}, 
 	Hui Zhang$^{\rm 4}$, 
        Yuan Zhang$^{\rm 1}$,
	Ajian Liu$^{\rm 5}$\thanks{Corresponding author.},
        Jun Wan$^{\rm 5}$,   
        Zhen Lei$^{\rm 5,6,7}$  \\
        $^{\rm 1}$China Telecom Research Institute (CTRI); 
        $^{\rm 2}$SIAT, Chinese Academy of Sciences; \\
        $^{\rm 3}$Southern University of Science and Technology; 
        $^{\rm 4}$Tianjin University of Science \& Technology; \\
	$^{\rm 5}$MAIS, CASIA, China;
 	$^{\rm 6}$SAI, UCAS, China; 
        $^{\rm 7}$CAIR, HKISI, CAS. \\
	\tt\footnotesize
	$^{\rm 1}$zouh3@chinatelecom.cn;
        $^{\rm 2}$cxdu025@163.com;
        $^{\rm 5}$ajianliu92@gmail.com
}
\maketitle
\thispagestyle{empty}


\begin{abstract}
Facial recognition systems are susceptible to both physical and digital attacks, posing significant security risks.
Traditional approaches often treat these two attack types separately due to their distinct characteristics. Thus, when being combined attacked, almost all methods could not deal.
Some studies attempt to combine the sparse data from both types of attacks into a single dataset and try to find a common feature space, which is often impractical due to the space is difficult to be found or even non-existent.
To overcome these challenges, we propose a novel approach that uses the sparse model to handle sparse data, utilizing different parameter groups to process distinct regions of the sparse feature space.
Specifically, we employ the Mixture of Experts (MoE) framework in our model, expert parameters are matched to tokens with varying weights during training and adaptively activated during testing. 
However, the traditional MoE struggles with the complex and irregular classification boundaries of this problem.
Thus, we introduce a flexible self-adapting weighting mechanism, enabling the model to better fit and adapt.
In this paper, we proposed La-SoftMoE CLIP, which allows for more flexible adaptation to the Unified Attack Detection (UAD) task, significantly enhancing the model's capability to handle diversity attacks.
Experiment results demonstrate that our proposed method has SOTA performance.
\end{abstract}

\section{Introduction}
Face recognition has been widely used in massive areas such as public security, monitoring identification, authentication, etc. 
It is always at risk of Physical Attacks (PAs) such as 3D masks, print attacks, and others. 
A series of competitions based on visual multimodality~\cite{zhang2020casia,liu2019multi}, multiethnicity~\cite{CeFAdatabase2021,liu2021cross}, high fidelity masks~\cite{liu2022contrastive,liu20213d}, surveillance scenes~\cite{fang2023surveillance,Fang_2023_CVPR}, and joint attack detection~\cite{fang2024unified,yuan2024unified} themes have attracted increasing attention from researchers.

Works of physical attack detection~\cite{yu2020face, liu2021face, liu2023fm, ijcai2022p165} always deal with this task by building special neural networks and integrating multi-modal data, such as RGB image, depth, rPPG, and so on~\cite{yu2022deep, yu2023visual, yu2019remote}.
With the development of generation models, Digital Attacks (DAs) seriously threaten the security of face recognition and become an important research area of Face Anti-Spoofing (FAS).
Digital attacks are more easily generated than physical ones by generative models such as DeepFack~\cite{ciftci2020fakecatcher}, StyleGAN~\cite{sarkar2022gan}, etc. 
Due to the large gap between these two types of attacks, the physical anti-spoofing methods could not handle this kind of soaringly increasing digital ones well. 
Thus, some works focus on digital attacks with methods based on CNN~\cite{lecun1998lenet} and Transformer~\cite{vaswani2017attention}, such as ~\cite{zhou2017two, korshunov2018deepfakes, rossler2019faceforensics++, dang2020detection}.

\begin{figure}[t!]
\centering
\includegraphics[width=3.3in]
{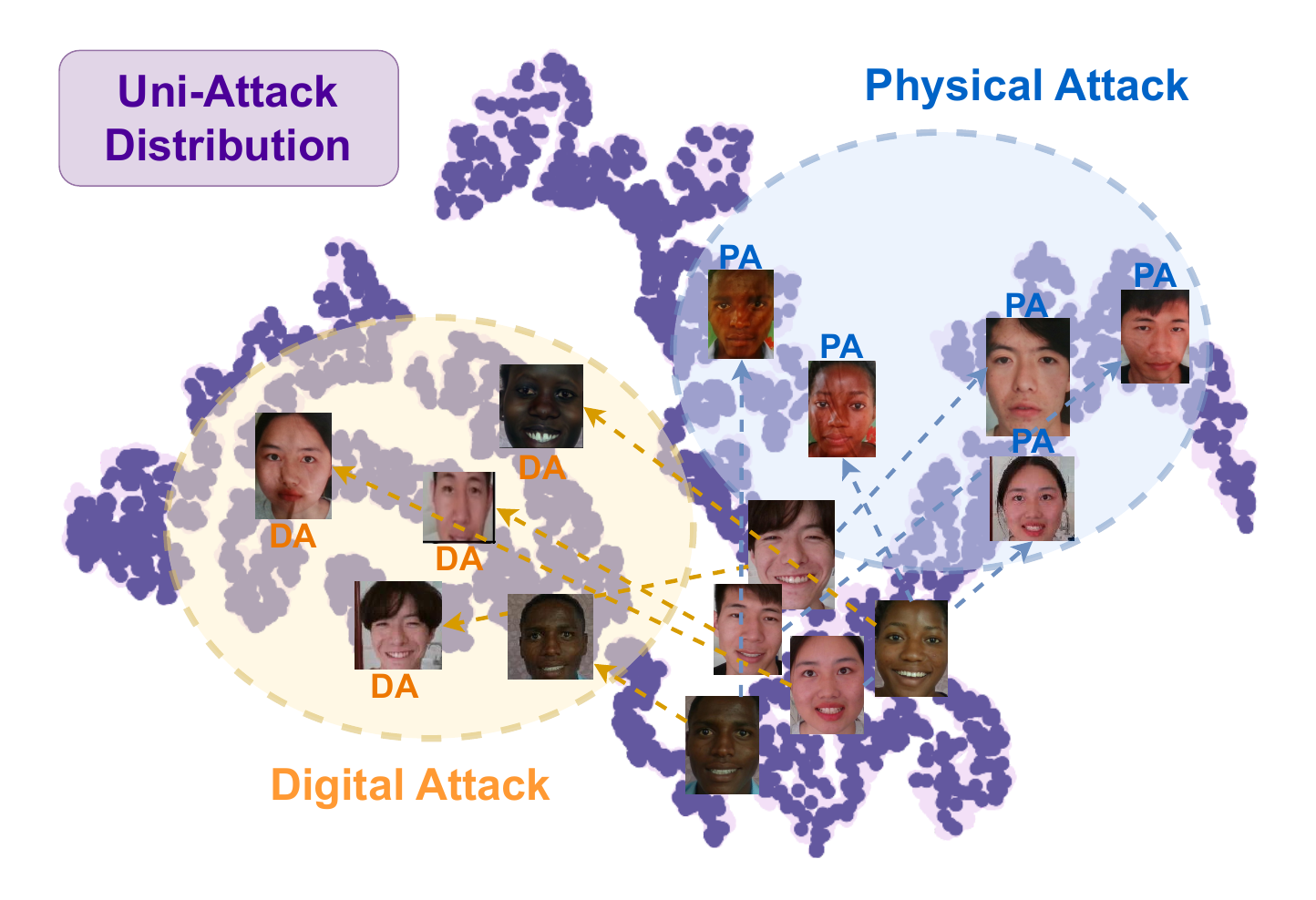}
\caption{
This figure shows the distribution of the UniAttackData~\cite{fang2024unified}. It has significant clustering and connectivity due to the ID consistency but has a big gap between PAs and DAs in the feature space.
}
\label{figure_distribution}
\hfil
\end{figure}

\begin{figure*}[ht!]
\centering
\includegraphics[width=6.5in]
{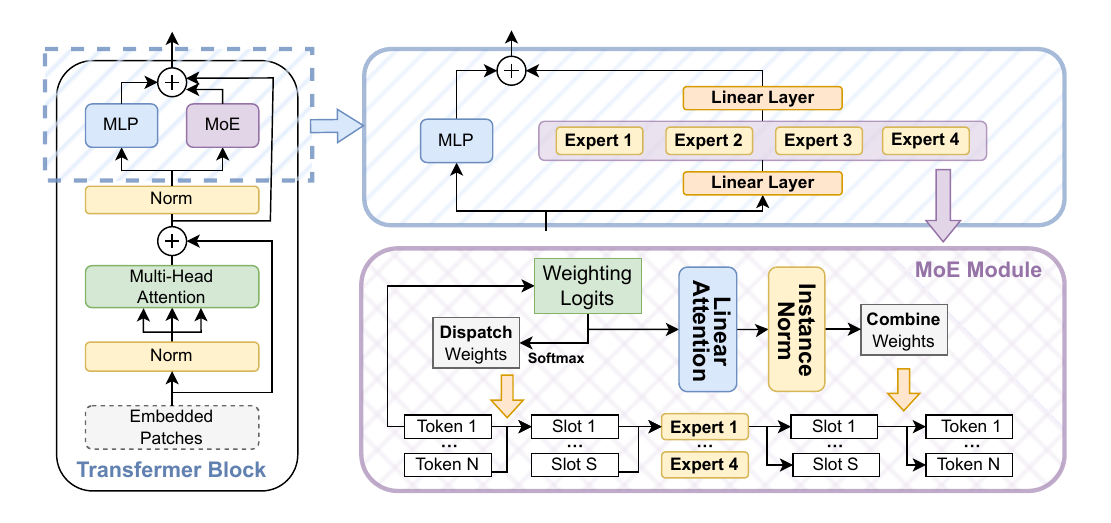}
\caption{
This figure shows the overall framework of our method. 
The left of the figure demonstrates the basic block structure of the image encoder of the CLIP~\cite{clip}, the whole image encoder consists of 12 Transformer Blocks.
Among that, the MoE Module is added in parallel with MLP and has two Linear layers at the input and output. 
Furthermore, we replaced Soft MoE's token querying mechanism with linear attention to allow for more flexible self-adapting to the sparse feature distribution of UAD. 
}
\label{figure_framework}
\hfil
\end{figure*}

Both physical and digital attacks could cause bad effects. 
Although current methods perform well on large datasets, single-model struggle to effectively address various types of attacks due to poor generalization.
Thus, according to the gap between these two types of attacks, previous works have dealt separately as independent topics: Physical Attack Detection (PAD) and Digital Attack Detection (DAD).
That significantly increased computational resource demands and inference time for deploying them respectively. 
So it is meaningful and urgent to deal with the physical and digital attack as a unified detection task.
In recent years, several works have appeared, Yu et al.~\cite{yu2024benchmarking} designs a dual branch physiological network, establishing the first joint face spoofing and forgery detection benchmark which uses visual appearance and physiological cues.
Fang and Liu et al.~\cite{fang2024unified} map physical and digital attacks into the same feature space, then use a Unified Knowledge Mining module to detect the fake faces with various spoofing types. 
It also proposes a dataset that guarantees ID consistency, which applies various physical and digital attacks on the face of the same ID.

Processing these two attacks with mapping into the same unified feature space achieves satisfactory results~\cite{yu2024benchmarking, fang2024unified}. 
These unified detecting methods could save the extra cost caused by separately training and deploying two models.
The weak generalization ability of previous models is commonly caused by the feature distribution gap between PAs and DAs. Unified detection could make the distribution of those two attacks aligned.
Despite these successes, inherent differences between physical and digital attacks remain. Both PAs and DAs are classified as “fake” in one solution space of face anti-spoofing detection, the significant differences between these two types of attacks increase the intra-class distance for the “fake” category. 
We demonstrate this in the Ablation Study section~\ref{Ablation Study}.
That may lead the models to overfit to a particular type of attack and not learn compact feature spaces for detecting both types of attacks, making challenges for models such as CLIP~\cite{clip} in fully utilizing its capabilities due to its limitations with sparse feature distribution in classification tasks.

CLIP (Contrastive Language–Image Pre-training)~\cite{clip} is a large visual-linguistic model based on contrastive learning~\cite{chen2020simpclr}, proposed by OpenAI. It consists of two components, an image encoder and a text encoder.
The main idea of CLIP is concise, and its structure is straightforward, which is designed to associate images with their corresponding textual descriptions to make up for the single mode. 
The image encoder, typically a ResNet~\cite{he2016deep} or ViT~\cite{ViT}, maps images into a feature space. While the text encoder, usually a Transformer, maps the textual descriptions of images into the same feature space. 
Then contrastive learning enables the model to grasp complex relationships between images and text, it maximizes the similarity between image-text pairs that describe the same content while minimizing the similarity between randomly paired images and texts.

CLIP has performed excellently in many visual tasks~\cite{yang2023continual,jiang2024effectiveness,zeng2024human,tang2024wfss,guo2023semantic}, for example, CFPL~\cite{liu2024cfpl} and FLIP~\cite{srivatsan2023flip}.
CFPL is built on CLIP and proposed by Liu et al. 
CLIP's powerful ability to deal with vision and language dual modalities provides a probability to use text features on FAS.
So, the CFPL uses dynamically adaptive text features as weights of the classifier to deal with Domain Generalization-based Face Aniti-Spoofing.
FLIP is a CLIP-based framework proposed by Srivatsan et al.
It contains three fine-tuned variants: FLIP-V, FLIP-IT, and FLIP-MCL. They are used to improve the generalization of models on cross-domain.

CLIP~\cite{clip} has great advantages in classifying dense feature distribution with its serial structure, but when faced with the problem of sparse feature space distribution, it performs weakly. 
So, in other words, CLIP's ability may not be fully utilized in resolving the problem of UAD because of the big gap between PAs and DAs in the feature space as shown in Figure~\ref{figure_distribution}.
Fortunately, Mixture of Experts (MoE), which has been widely used in large-scale language generation models in recent years, provides an effective resolution to resolve the problem.
It is a sparse gate-controlled deep learning model consisting of a set of experts and a gating network~\cite{fedus2022review}, that divides the task into several sub-tasks, coped by different experts.
This mechanism allows dense models with MoE to perform better on data with sparse distribution. Because each expert just concentrates and deals with distinct features or sub-tasks when both training and inference processes, the performance of models could be improved integrally without obviously increasing time cost.

In this work, we employ SoftMoE~\cite{puigcerver2023sparse} with CLIP to improve the model’s performance on UAD. 
More specifically, we introduce the SoftMoE in CLIP’s image encoder to make it more suitable for the sparse feature distribution of UAD. Moreover, we replaced the weighting method of SoftMoE with Linear Attention~\cite{linearattention} and named the new method La-SoftMoE. Section~\ref{Method} shows the details of our methods.
The experimental result proves that our proposed methods could effectively improve the detection performance of UAD tasks as shown in Section~\ref{Experiments}. The main contributions are as follows:
\begin{itemize}
    \item 
    We introduce the SoftMoE into the image encoder of CLIP to improve its ability to deal with sparsity data distribution of UAD tasks.
    \item 
    We proposed La-SoftMoE which replaces the weighting method of SoftMoE with Linear Attention. And evaluate the La-SoftMoE CLIP on the UAD tasks to prove its performance.
\end{itemize}

\section{Related Work}

\subsection{Unified Attack Detection (UAD)}
To detect the PAs and DAs in face recognition systems, researchers provided lots of effective methods in previous works. For example, 
CA-MoEiT~\cite{liu2024moeit} improves the generalization of the model in terms of feature enhancement, feature alignment and feature complementation through the introduction of MixStyle, Double Cross Attention Mechanism and Semi-fixed Mixture of Experts (SFMoE) in the Visual Transformer (ViT).
Rossler et al.~\cite{rossler2019faceforensics++} proposed a method that extracted features from facial images, conducting binary classification (“real” or “fake”) through CNN to determine whether the detected sample is a real face.
Zhao et al.~\cite{zhao2021multi} researched how to identify more difficult-to-identify fake patterns in fake faces, such as noise statistics, local textures, and frequency information.
Deb et al.~\cite{deb2023unified} introduced a unified detection idea that used the k-means algorithm to cluster 25 attack methods in the first stage. Then a Multi-task Learning framework is conducted to distinguish real faces from fake ones.
Most previous works use the dataset without ID consistency, which would cause models to learn non-critical factors such as background, ID of samples, and other bio-irrelevant information.
UniAttackData~\cite{fang2024unified} is a UAD dataset with ID consistency which was published in 2024 by Fang and Liu et al., and the authors also designed a Unified Knowledge Mining module based on this dataset to learn the knowledge of physical-digital joint feature space. Its experiment result confirms that this dataset with ID consistency could improve the detection ability of models on UAD tasks.

\begin{figure*}[ht!]
\centering
\includegraphics[width=6.5in]
{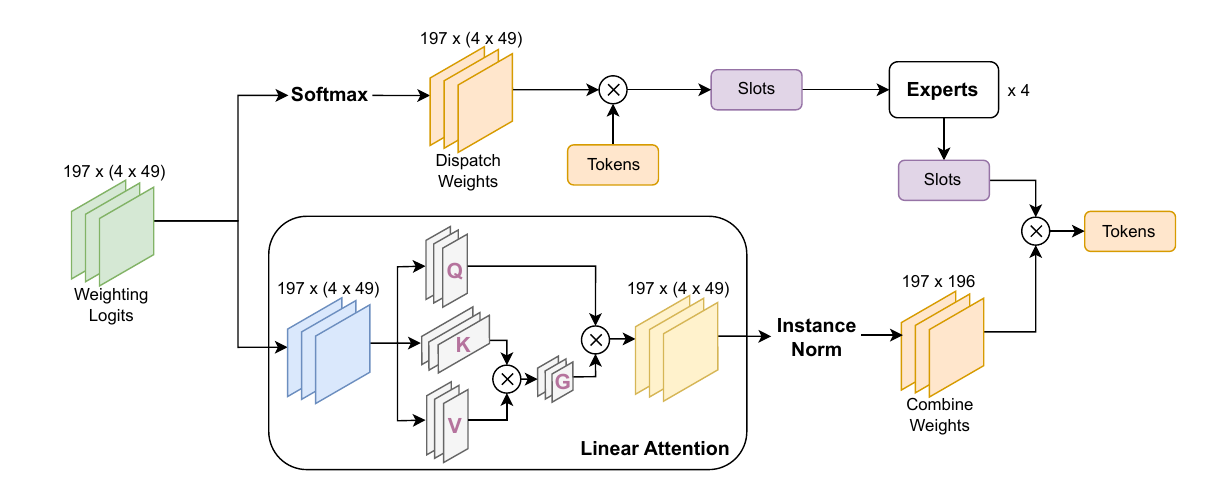}
\caption{This figure shows a part of the dimension transformation of La-SoftMoE. The number of experts is 4, of slots 49, and features 768. Dispatch Weights maps the input tokens into slots for different experts by weighting all tokens. Combine Weights query the corresponding tokens of slots. We improve the query manner of output tokens by replacing the softmax of combined weights with linear attention and an Instance Norm layer. The Instance Norm maps the weights into the range (0,1).}
\label{figure_exchange}
\hfil
\end{figure*}

\subsection{Multimodal Large Language Models (MLLMs)}
With the rise of Large Language Models (LLMs) in recent years, incorporating multi-modal information into FAS tasks, particularly text descriptions, has become feasible, which significantly enhances the models' performance~\cite{yin2023survey}.
LLMs could learn knowledge from cross-modalities. They are usually used to match samples, such as images, with their corresponding descriptions.
Thus, many works employed LLM to introduce multi-modal information to get better performance. For example, 
M-ICL (Multimodal In-Context Learning)~\cite{yin2023survey} extends the concept of traditional in-context learning, which allows models to learn multi-modal in-context.
Li et al.~\cite{li2023mimic} built a dataset that combines multi-modal context and instruction tuning to improve model performance in image description tasks.
M-CoT (Multimodal Chain of Thought)~\cite{yin2023survey} applies the CoT (Chain of Thought) reasoning strategy to multi-modal tasks.
Zhang et al.~\cite{zhang2023multimodal} used the ScienceQA dataset to reason the steps.
The model first generates intermediate reasoning steps (thinking chains) and then derives the final answer based on these steps.

\subsection{Mixture-of-Experts (MoEs)}
The development of Mixture-of-Experts (MoEs) has a history of at least 30 years~\cite{fedus2022review}.
The first appearance of the MoEs was in 1991, proposed by Jacobs et al.~\cite{jacobs1991adaptive}.
It is a supervised learning process, adaptive mixtures of local experts, which is a system that contains multiple separate networks processing subsets of all training samples. 
Recently, the widely used idea of MoEs was proposed by Noam et al.~\cite{shazeer2017outrageously} in the Sparsely-Gated Mixture-of-Experts Layer, which contributed sparsely-gated mechanism and token-level process.
With the development of LLMs, there is an urgent need for a method that can improve the performance of the model on sparse tasks.
GShard~\cite{lepikhin2020gshard} provides an efficient way, which is the first work to apply the ideas of MoE on a Transformer.
Zoph et al. proposed ST-MoE~\cite{zoph2022st} to resolve training instabilities and uncertain quality problems during fine-tuning in language models.
In the previous MoE, each expert just got several tokens and could not get information on other tokens which would cause two common problems, token dropping and expert imbalance problems.
Different from them, each slot in Soft MoE~\cite{puigcerver2023sparse} is a weighted average of all tokens to avoid common problems. 
In addition, benefiting from two weighting operations, Soft MoE completely avoids sorting the top-k experts, which makes it faster than most previous sparse MoEs.

\section{Method}\label{Method}
In this work, we propose a framework based on CLIP~\cite{clip} to deal with UAD tasks. 
However, the feature distribution of the UAD task is sparse, so the capability of CLIP may not be fully utilized.
To further improve the performance of CLIP on dealing with UAD tasks, we introduce Soft MoE~\cite{puigcerver2023sparse}, which has achieved satisfactory success on tasks with sparse feature distribution in previous works.
Moreover, we replace the token query manner in Soft MoE with linear attention, making it more suitable for our task.

Because of CLIP's powerful performance on image classification tasks, we employ it in dealing with UAD in this work, which focuses on identifying “fake” human faces from “live” human faces.
CLIP has two encoders: an image encoder and a text encoder. 
The image encoder is usually composed of ViT~\cite{ViT} or ResNet~\cite{he2016deep} and in this work, we apply ViT with Soft MoE in each block as the backbone of it.
The input of the image encoder in both training and inference processes is the UniAttackDate~\cite{fang2024unified} which includes live faces and their corresponding fake faces with PAs and DAs.
The input of the text encoder is a sentence with the label {fake/live}. 
And most of the experiments in our work, the sentence is “This is an image of a \textless{fake/live}\textgreater face”.

In our model, we keep the loss function of the CLIP model, the cross entropy of the similarity between images and texts. The formula is shown as Equation\ref{CLIPLOSS}, where $\mathbf{S}$ is the similarity matrix, and the element $S_{ij}$ of the similarity matrix $\mathbf{S}$ denotes the similarity between the $i^{th}$ image embedding and the $j^{th}$ text embedding. 

\begin{small} 
\begin{equation}
    \begin{aligned}
    \label{CLIPLOSS}
    L=-\frac{1}{2N}\sum_{i=1}^N\left(\log\frac{\exp(S_{ii})}{\sum_{j=1}^N\exp(S_{ij})}+\log\frac{\exp(S_{ii})}{\sum_{j=1}^N\exp(S_{ji})}\right)
    \end{aligned}
\end{equation}
\end{small} 

We add Soft MoE to the image encoder to improve the performance of CLIP on dealing with UAD, the task with sparse distribution.
Previous MoEs assign several tokens for one expert and select top-k experts.
This mechanism usually causes problems such as token dropping, extensive computing costs, etc.
Unlike in previous MoEs, experts in Soft MoE gain several weighted averages of all features, called slots, instead of discrete distinct tokens.
This characteristic of Soft MoE transforms the discrete optimization problem of sparse MoE Transformers into a differentiable optimization problem and alleviates the token-dropping problem.
So, in Soft MoE, it is not necessary to find a one-to-one match between the input token and the experts “hardly”, but could mix the input tokens “softly” and assign them to each expert.
Another characteristic of Soft MoE is avoiding to select top-k experts for each input token.
Thanks to it, the computing cost of sorting and determining experts can be reduced and it becomes easier to extend the number of experts to even thousands.

In Soft MoE, the tokens are weighted by dispatch weights $\mathbf{D}$ and mapped to slots. 
This process allows experts to get a set of weighted averages of tokens as inputs instead of individual tokens and alleviates token-dropping issues. 
After the slots pass through experts, the slots are then weighted by combined weights $\mathbf{C}$ to query the output tokens. 
This process is employed to query the output token list of Soft MoE.
The computing process is shown as \ref{dispatch_weights}, \ref{experts}, \ref{combine_weights}.

\begin{equation}
    \begin{aligned}
    \label{dispatch_weights}
    \mathbf{D}_{ij}=\frac{\exp((\mathbf{X}\boldsymbol{\Phi})_{ij})}{\sum_{{i^{\prime}=1}}^{n}\exp((\mathbf{X}\boldsymbol{\Phi})_{{i^{\prime}j}})},\quad&\tilde{\mathbf{X}}=\mathbf{D}^{\top}\mathbf{X}
    \end{aligned}
\end{equation}

$\mathbf{X}\in\mathbb{R}^{{n\times d}}$ denotes the input tokens of Soft MoE, where $\textit{n}$ is number of tokens and $\textit{d}$ is their dimension.
In Soft MoE with $\textit{e}$ experts, each expert deals with $\textit{s}$ slots where per slot has a corresponding vector of parameters with $\textit{d}$-dimension. 
The corresponding vectors are denoted by $\mathbf{\Phi}\in\mathbb{R}^{{d\times(e\cdot s)}}$. 
$\mathbf{D}\in\mathbb{R}^{{n\times(e\cdot s)}}$ is the dispatched weights, which is the result of applying a softmax over the columns of $\mathbf{X}\mathbf{\Phi}$.
${D}_{ij}$ denotes the element in dispatched weights ${D}$ located in the ${i}^{th}$ column and ${j}^{th}$ row.
$\tilde{\mathbf{X}}\in\mathbb{R}^{(e\cdot s)\times d}$ is the input of Experts, a set of slots. It is obtained by weighting $\mathbf{X}$ with the dispatch weights $\mathbf{D}$.

\begin{equation}
    \begin{aligned}
    \label{experts}
    \tilde{\mathbf{Y}}=Experts(\tilde{\mathbf{X}})
    \end{aligned}
\end{equation}

$\tilde{Y}=\{\tilde{Y}_{i}|i\in\{1,2,\cdots,e\}\}$ is the output slots, where $\tilde{\mathbf{Y}}_i=expert_i(\tilde{\mathbf{X}}_i)$ denotes the output of each expert.

\begin{equation}
    \begin{aligned}
    \label{combine_weights}
    \mathbf{C}_{ij}=\frac{\exp((\mathbf{X}\boldsymbol{\Phi})_{ij})}{\sum_{j^{\prime}=1}^{e\cdot s}\exp((\mathbf{X}\boldsymbol{\Phi})_{ij^{\prime}})},\quad\mathbf{Y}=\mathbf{C}\tilde{\mathbf{Y}}
    \end{aligned}
\end{equation}

$\mathbf{C}\in\mathbb{R}^{{n\times(e\cdot s)}}$ is the combined weights, which is the result of applying a softmax over the rows of $\mathbf{X}\mathbf{\Phi}$. 
${C}_{ij}$ denotes the element in combined weights ${C}$ located in the ${i}^{th}$ column and ${j}^{th}$ row.
$\mathbf{Y}$ is the output tokens of Soft MoE.

The image encoder of our model is composed of ViT, which consists of 12 residual attention blocks.
The Soft MoE is added as parallel with MLP in each residual block, as shown in Figure \ref{figure_framework}.
The MoE Module of Figure \ref{figure_framework} shows La-SoftMoE, which is a modified Soft MoE. 
Compared to the original Soft MoE, La-SoftMoE has a linear attention mechanism and an instance normalization layer between weighting logits and combined weights $\mathbf{C}$.
The linear attention is used to modify the weighting manner of combined weights $\mathbf{C}$.
The instance normalization layer is used to keep each weight in $\mathbf{C}$ is non-negative.
It has been shown in Figure~\ref{figure_exchange} that the dimension exchanging process of data when crossing La-SoftMoE.

The combined weights $\mathbf{C}$ in the original Soft MoE are used to query the corresponding tokens of each slot from weighting logits($\mathbf{X}\mathbf{\Phi}$) and it is obtained by applying a softmax on weighting logits over its rows.
This process makes the weighting manner on each slot the same. 
But slots are the output of different experts, which might be distinct.
So it's necessary to use a more reasonable and special way for each of them to be weighted.

\begin{algorithm}[t]
    \caption{$Image Encoder$} 
    \label{ImageEncoder}
    \begin{algorithmic}[1] 
        \Procedure{imageencoder}{$I$}       
        \State $i \gets 0$
        \State $X \gets I$
        \While{$i < 12$}              
        \State $Y \gets Layernorm(X)$ 
        \State $X \gets X + Att(Y)$ 
        \State $X\_MLP \gets mlp(X)$
        \State $X\_MoE \gets La-SoftMoE(X, 4, 49)$
        \State $X \gets X + X\_MLP + X\_MoE$ 
        \State $i \gets i + 1$
        \EndWhile
        \State $I\_f \gets X$
        \State \textbf{return} $I\_f$  
        \EndProcedure
    \end{algorithmic}
\end{algorithm}

\begin{algorithm}[t]
    \caption{$La-SoftMoE$} 
    \label{SoftMoE}
    \begin{algorithmic}[1] 
        \Procedure{softmoelayer}{$X,\ e,\ s$}       
        \State $n, d \gets X.shape$
        \State $\Phi \gets randn(e,\ s,\ d)$
        \State $logit \gets einsum(n\ d,\ e\ s\ d ->\ n\ e\ s,\ X,\ \Phi)$      
        \State $D \gets logits.softmax(dim = 0)$ 
        \State $C \gets rearrange(logits, \  n\ e\ s\ ->\ n\ (e\ s))$
        \State $C \gets LinearAttention(C)$
        \State $C \gets InstanceNorm(C)$
        \State $slots \gets einsum(n\ d,\ n\ e\ s\ ->\ e\ s\ d,\ X,\ D)$
        \State $out \gets experts(slots)$
        \State $out \gets rearrange(out,\ e\ s\ d\ ->\ (e\ s)\ d)$
        \State $out \gets einsum(s\ d,\ n\ s\ ->\ b\ n\ d,\ out,\ C)$
        \State \textbf{return} $out$  
        \EndProcedure
    \end{algorithmic}
\end{algorithm}

The fundamental operation of the attention mechanism is to calculate the attention weights. 
These weights are determined by computing the similarity between the query and each key. 

In conventional dot-product attention~\cite{vaswani2017attention}, the computation of attention weights is shown as \ref{att}, where $\mathbf{Q}$ is the query matrix, $\mathbf{K}$ is the key matrix, $\mathbf{V}$ is the value matrix, and $d_{k}$ is the dimension of the keys.

\begin{equation}
    \begin{aligned}
    \label{att}
    \textit{Attention}(Q,K,V)=\textit{softmax}\left(\frac{QK^T}{\sqrt{d_k}}\right)V
    \end{aligned}
\end{equation}

The time complexity of dot product attention increases extensive computational cost. Many works aim at reducing computing costs through different perspectives. 
Wang et al.~\cite{wang2020linformer} demonstrated that the context mapping matric of self-atttion is a low-rank matrix and approximated self-attention by a low-rank matrix.
Shen et al.~\cite{shen2021efficient} exchanged the order of dot products of query, key, and value in self-attention. The softmax is replaced with a linear kernel mapping function in their proposed effective attention.
Katharopoulos et al.~\cite{linearattention} presented a proof of linear attention and applied it to their Fast Autoregressive Transformer.
The linear attention is play and plug.
Similar to effective attention, it removes the softmax operation used in traditional attention and introduces a feature mapping function $\phi$ to transform the input query and key matrices into a new feature space instead. 
The formula for linear attention is shown in Equation \ref{linearAtt}.

\begin{equation}
    \begin{aligned}
    \label{linearAtt}
    \textit{LinearAtt}(Q,K,V)=\phi(Q)\begin{pmatrix}\phi(K)^TV\end{pmatrix}
    \end{aligned}
\end{equation}

In the MoE Module part of Figure \ref{figure_framework}, it can be noticed that the combined weights $\mathbf{C}$ provide a way for slots to query the corresponding tokens, which is similar to the attention mechanism. 
So we replace the weighting manner of slots with attention to query output tokens.
Considering the time and memory complexity of dot-product self-attention mechanisms, we decided to use linear attention with linear computational cost in this fine-tuning phase to ensure that the computational cost of the model does not increase significantly.

The formula after fine-tuning is shown as Equation \ref{la-softmoe}.

\begin{equation}
    \begin{aligned}
    \label{la-softmoe}
    \mathbf{C}=LinearAtt(\mathbf{X}\boldsymbol{\Phi}),\quad\mathbf{Y}=\mathbf{C}\tilde{\mathbf{Y}}
    \end{aligned}
\end{equation}

$LinearAtt(\mathbf{X}\boldsymbol{\Phi})$ denotes using Linear Attention to focus the key area of each slot and then obtain weighting metric.
We refer to this fine-tuned SoftMoE as La-SoftMoE.

Algorithm \ref{SoftMoE} shows the working process of La-SoftMoE. 
In the pseudo-code, $\textit{einsum}$ refers to the Einstein summation convention, which is used for matrix multiplication here. 
Algorithm \ref{ImageEncoder} details the workflow of our modified image encoder. 
The image encoder contains 12 residual attention blocks, integrated with a La-SoftMoE with 4 experts and 49 slots into each residual block, parallel with the MLP.


\section{Experiments}\label{Experiments}

\subsection{Datasets and Experimental Setting}

In this paper, we employed UniAttackData~\cite{fang2024unified} and JFSFDB~\cite{yu2024benchmarking} as experimental datasets. 
The UniAttackData is the primary evaluation dataset due to the ID consistency. Moreover, to prove the effectiveness of our proposed methods, we verified the evaluation results with different prompts. 
The ablation study is also proposed in Section~\ref{Ablation Study}.

The UniAttackData~\cite{fang2024unified} is a unified physical-digital attack dataset proposed by Fang and Liu et al.. It is ID consistent with 2 types of PAs and 12 types of DAs involving 1800 subjects.
More specifically, the types of DAs include 6 editing and 6 adversarial,  which is the most advanced and comprehensive attack method to our knowledge.
Table~\ref{dataset} shows the scale of this dataset, which includes 600 subjects / 8400 images for training, 300 subjects / 6000 images for validation, and 900 subjects / 21506 images for testing.
Two protocols are defined in this dataset, first one applies live faces and all their corresponding attacks on training, validation, and test subsets, while the other one is used to evaluate the model's generalization on unseen samples. 
In this work, we only use the first protocol to evaluate our proposed methods.
The JFSFDB~\cite{yu2024benchmarking} is proposed by Yu et al with the combining of nine subsets, including SiW~\cite{liu2018SiW}, 3DMAD~\cite{erdogmus20143DMAD}, HKBU-MarsV2 (HKBU)~\cite{liu20163d}, MSU-MFSD (MSU)~\cite{wen2015face}, 3DMask~\cite{yu2020fas} and ROSE-Youtu (ROSE)~\cite{li2018unsupervised} for PAD while FaceForensics++ (FF++)~\cite{rossler2019faceforensics++}, DFDC~\cite{dolhansky2019deepfake} and CelebDFv2~\cite{li2020celeb} for DAD.
It has two main protocols: separate training and joint training, wherein the separate training allows models to deal with PA and DA tasks respectively.
we choose the joint training protocol to evaluate our method in Section~\ref{Additional Analysis}.

\begin{figure}[t]
\centering
\includegraphics[width=3.4in]
{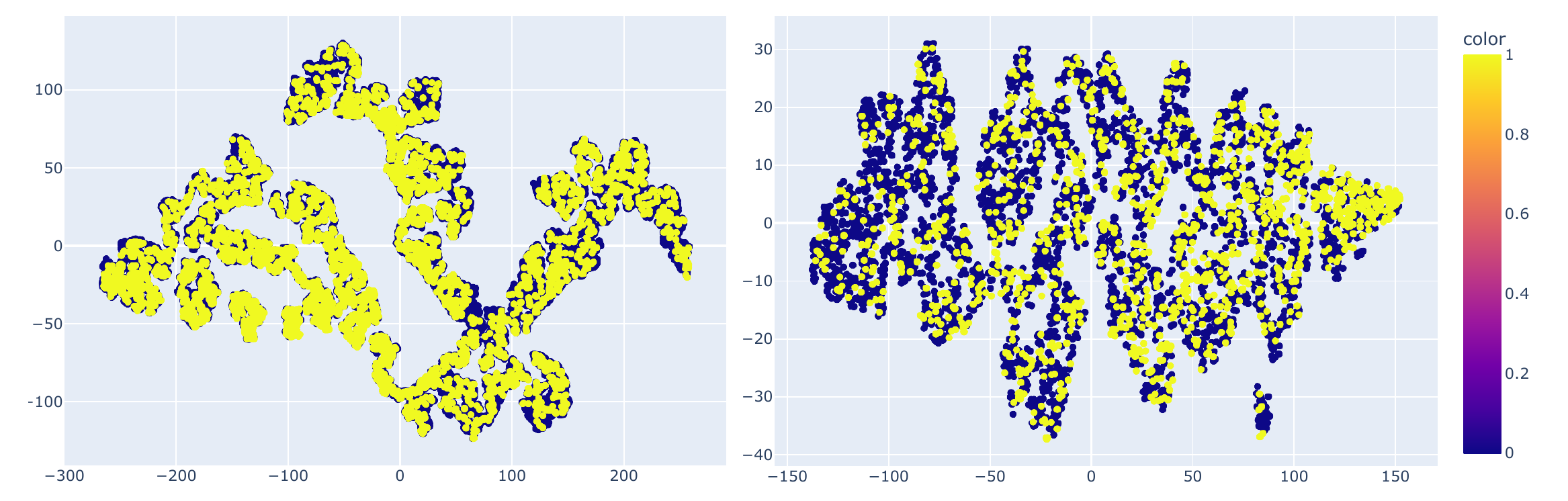}
\caption{This figure shows the distribution of the UniAttackData~\cite{fang2024unified} (left) and the JFSFDB~\cite{yu2024benchmarking} (right).  UniAttackData has significant clustering and connectivity, while JFSFDB is more scattered with no significant clustering trend.}
\label{figure_dis-datasets}
\hfil
\end{figure}

 \begin{table}\scriptsize

     \centering
     \resizebox{\linewidth}{!}{
     \begin{tabular}{c|c|cccc|c}
         \hline
          Class    &\makecell{Num of \\Subjects} & Live & Phys & Adv & Digital & Total\\
         \hline
         train	&600 &3000	&1800	&1800	&1800  &8400  \\
         eval	&300 &1500	&900	&1800	&1800   &6000  \\
         test	&900 &4500	&2700	&7106	&7200   &21506  \\
         \hline
     \end{tabular}}
     \caption{This table shows the scale of the UniAttackData~\cite{fang2024unified}. Including the number of train/eval/test subjects and images of each type.}
     \label{dataset}
 \end{table}

 \begin{table}
     \centering
     \resizebox{\linewidth}{!}{
     \begin{tabular}{c|cccc}
         \hline
          Method	&ACER(\%)$\downarrow$	&ACC(\%)$\uparrow$	&AUC(\%)$\uparrow$	&EER(\%)$\downarrow$ \\
         \hline
         ResNet50~\cite{lecun1998lenet}	&1.35	&98.83	&99.79	&1.18\\
         VIT-B/16~\cite{ViT}	&5.92	&92.29	&97.00	    &9.14\\
         Auxiliary~\cite{liu2018learning}	&1.13	&98.68	&99.82	&1.23\\
         CDCN~\cite{yu2020searching}	    &1.4	&98.57	&99.52	&1.42\\
         FFD~\cite{dang2020detection}	        &2.01	&97.97	&99.57	&2.01\\
         UniAttackDetection~\cite{fang2024unified}&0.52   &99.45  &\textbf{99.95} &\textbf{0.53}\\
        \hline
         \textbf{Ours}	    &\textbf{0.32}	&\textbf{99.54}	&99.72	&0.56\\

         \hline
     \end{tabular}}
     \caption{
     This table shows the results with the UniAttackData~\cite{fang2024unified}, our proposed La-SoftMoE CLIP has the best ACER 0.32\% and ACC 99.54\%, achieves the SOTA performance.
     }
     \label{result_exp}
 \end{table}

The experimental setup details are shown below: The backbone of the image encoder in CLIP is ViT-B/16~\cite{ViT}. We choose 4 experts for each Transformer Block, and the slot number is set to 49. The training processing is running on the NVIDIA A100 GPU with the Adam optimizer and the learning rate is $10^{-6}$.

\subsection{Performance Evaluation}

To evaluate the performance of our proposed method La-SoftMoE CLIP, we choose the Average Classification Error Rate (ACER), Overall Detection Accuracy (ACC), Area Under the Curve (AUC), and Equivalent Error Rate (EER) as the performance evaluation metrics.  
And compare our methods with ResNet50~\cite{lecun1998lenet}, ViT-B/16~\cite{ViT}, FFD~\cite{dang2020detection}, CDCN~\cite{yu2020searching}, Auxiliary~\cite{liu2018learning}, and UniAttackDetection~\cite{fang2024unified}. 
Table~\ref{result_exp} shows the results with the UniAttackData~\cite{fang2024unified}, our proposed method has the best ACER and ACC, which means it effectively filters out spoofing ones and achieves the SOTA performance.
More specifically, our ACER is 0.32\% which is far better than the best previous (0.52\%). The ACC achieves 99.54\% which is better than the best previous (99.45\%). 
The AUC and EER are slightly worse than the best previous results, our AUC is 99.72\%, compared to 99.95\%, and our EER is 0.56\%, compared to 0.53\%.

 \begin{table}
     \centering
     \resizebox{\linewidth}{!}{
     \begin{tabular}{c|cccc}
         \hline
          Method	&ACER(\%)$\downarrow$	&ACC(\%)$\uparrow$	&AUC(\%)$\uparrow$	&EER(\%)$\downarrow$ \\
         \hline
         ResNet50~\cite{lecun1998lenet}	&7.70	&90.43	&98.04	&6.71\\
         VIT-B/16~\cite{ViT}	&8.75	&90.11	&98.16	    &7.54\\
         Auxiliary~\cite{liu2018learning}	&11.16	&87.40	&97.39	&9.16\\
         CDCN~\cite{yu2020searching}	    &12.31	&86.18	&95.93	&10.29\\
         FFD~\cite{dang2020detection}        &9.86	&89.41	&95.48	&9.98\\
         \hline
         \textbf{Ours}	    &\textbf{4.21}	&\textbf{95.85}	&\textbf{99.11}	&\textbf{4.19}\\

         \hline
     \end{tabular}}
     \caption{
     This table shows the results with the JFSFDB~\cite{yu2024benchmarking}, our proposed La-SoftMoE CLIP has the SOTA performance. 
     }
     \label{result_JF}
 \end{table}

\subsection{Additional Analysis}\label{Additional Analysis}

To further evaluate the performance of our method La-SoftMoE CLIP, we also tested it on the JFSFDB~\cite{yu2024benchmarking}.
Table~\ref{result_JF} shows the results of our method and previous works, it has the SOTA performance with ACER 4.21\%, ACC 95.85\%, AUC 99.11\% and EER 4.19\%.
Though it has good performance, its evaluation results have gaps with the results of the UniAttackData~\cite{fang2024unified}, as shown in Table~\ref{result_exp}.
It may be caused by the differences between the two datasets UniAttackData~\cite{fang2024unified} and the JFSFDB~\cite{yu2024benchmarking}, that UniAttackData is ID consistent which means each subject has both PAs and DAs samples.
Figure~\ref{figure_dis-datasets} shows the distributions of both these two datasets by t-SNE~\cite{van2008visualizing}, the left one is UniAttackData while the right one is JFSFDB. 
The UniAttackData data points are primarily concentrated in the central region of the axis and extend along irregular paths, exhibiting significant clustering and connectivity. 
While the JFSFDB data points are more dispersed across the entire axis and show a relatively uniform distribution pattern, more scattered, with no significant clustering trend.

\begin{figure}[t!]
\centering
\includegraphics[width=3.3in]
{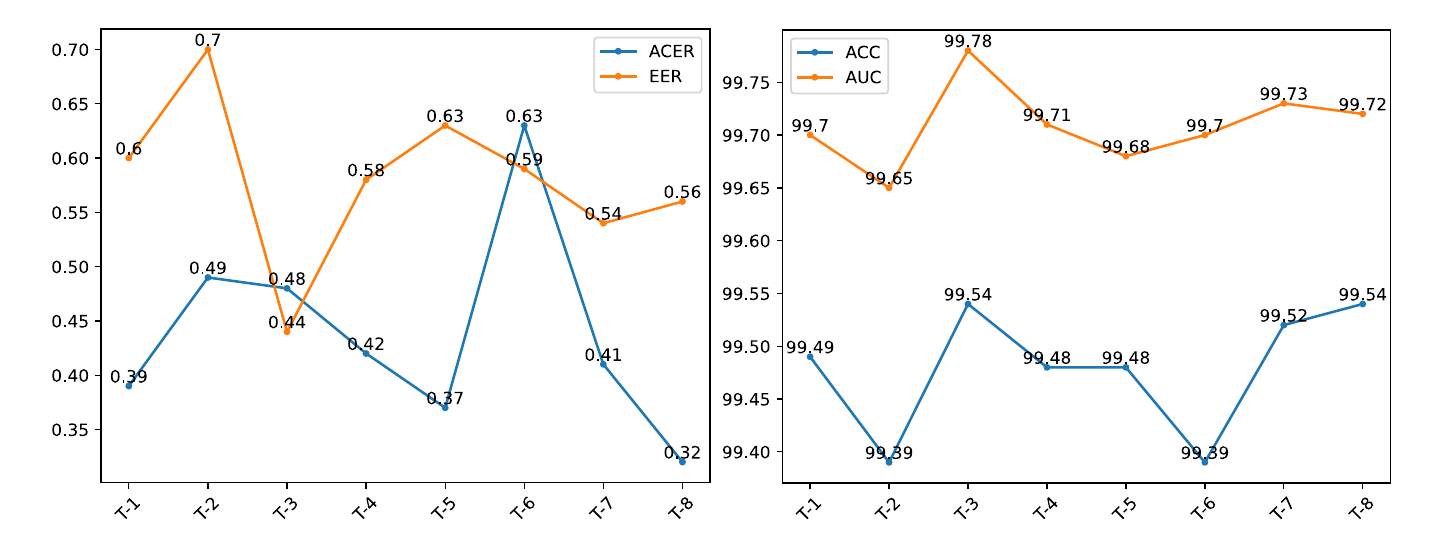}
\caption{
The performance of our model on the selected prompts. Their corresponding context is shown in Table~\ref{setting_prompts}. The left part shows the ACER and EER of using these sentences while the right part shows the ACC and AUC.
}
\label{figure_text}
\hfil
\end{figure}

Recently, many works~\cite{zhou2022learning, zhou2022conditional} have shown that prompts have a great impact on the performance of large models such as CLIP.
Thus, to prove our proposed method's effectiveness, we verified the evaluations of La-SoftMoE CLIP with different prompts.
As Table~\ref{setting_prompts} shows, We set 8 different sentences T-1 $\sim$ T-8 with the label “fake” or “live” as prompts. They describe the same meaning with different vocabulary and grammatical structures.
The evaluation results of each prompt with UniAttackData are shown in Table~\ref{result_prompts} and Figure~\ref{figure_text}. T-3 has the best ACC, AUC, and EER, while T-8 has the best ACER and ACC.
And though the prompts really affect large, all of the results show good performance, which proves that our proposed La-SoftMoE is effective and stable.
We choose the T-8 as the prompt of La-SoftMoE in other experiments part because of the outstanding performance of the ACER. It has the SOTA results of ACER and ACC as shown in Table~\ref{result_exp}. 
And addition, the T-3 has the SOTA results of ACC and EER.

 \begin{table}
     \centering
     \resizebox{\linewidth}{!}{
     \begin{tabular}{c|c}
         \hline
          Prompts	&Sentences \\
         \hline
         T-1	&There is a \textless{fake/live}\textgreater face in this photo.\\
         T-2	&\textless{fake/live}\textgreater face is in this photo.\\
         T-3	&A photo of a \textless{fake/live}\textgreater face.\\
         T-4	&This is an example of a \textless{fake/live}\textgreater face.\\
         T-5  &This is how a \textless{fake/live}\textgreater face looks like.\\
         T-6  &This photo contains \textless{fake/live}\textgreater face.\\
         T-7	&The picture is a \textless{fake/live}\textgreater face.\\
         T-8  &This is an image of a \textless{fake/live}\textgreater face.\\
         \hline
     \end{tabular}}
     \caption{
     This table shows the 8 prompt sentences T-1 $\sim$ T-8 we used. They describe the same meaning with different vocabulary and grammatical structures, and the \textless{fake/live}\textgreater is the label.
     }
     \label{setting_prompts}
 \end{table}

 \begin{table}
     \centering
     \resizebox{\linewidth}{!}{
     \begin{tabular}{c|cccc}
         \hline
          Prompts	&ACER(\%)$\downarrow$	&ACC(\%)$\uparrow$	&AUC(\%)$\uparrow$	&EER(\%)$\downarrow$ \\
         \hline
         T-1	&0.39	&99.49	&99.70	&0.60\\
         T-2	&0.49	&99.39	&99.65	&0.70\\
         T-3	&0.48	&\textbf{99.54}	&\textbf{99.78}	&\textbf{0.44}\\
         T-4	&0.42	&99.48	&99.71	&0.58\\
         T-5  &0.37	&99.48	&99.68	&0.63\\
         T-6  &0.63	&99.39	&99.70	&0.59\\
         T-7	&0.41	&99.52	&99.73	&0.54\\
         T-8  &\textbf{0.32}	&\textbf{99.54}	&99.72	&0.56\\
         \hline
     \end{tabular}}
     \caption{
     This table shows the La-SoftMoE CLIP's evaluation results with T-1 $\sim$ T-8 prompts corresponding to Table~\ref{setting_prompts}. 
     T-3 has the best ACC, AUC, and EER, while T-8 has the best ACER and ACC.
     }
     \label{result_prompts}
 \end{table}

\subsection{Ablation Study}\label{Ablation Study}

To verify the effectiveness of our contributions, we make the ablation experiments on vanilla CLIP, CLIP + Soft MoE, and CLIP + La-SoftMoE. Choose the UniAttackData~\cite{fang2024unified} as the dataset.
As shown in Table~\ref{result_abl}, after introducing the Soft MoE, the performance has been significantly improved, with the ACER down to 0.53\% which is far better than the vanilla CLIP (0.91\%), ACC up to 99.39\% better than the previous (98.87\%), and EER down to 0.68\% far better than the previous (0.96\%).
The AUC is slightly worse than the vanilla CLIP results, which is 99.66\% compared to 99.76\%.
It proves that MoE could help improve the classification performance of CLIP on sparse distribution tasks like UniAttackData.
After replacing Soft MoE with La-SoftMoE, the performance further improved, with the ACER down to 0.32\% which is far better than the CLIP + Soft MoE (0.53\%), ACC up to 99.54\% better than the previous (99.39\%), and EER down to 0.56\% better than the previous (0.68\%).
The AUC is better than the CLIP + Soft MoE but still slightly worse than the vanilla CLIP results, which is 99.72\%.
It proves that linear attention improves the query manner of Soft MoE so that could perform better.

\begin{figure}[t]
\centering
\includegraphics[width=3.4in]
{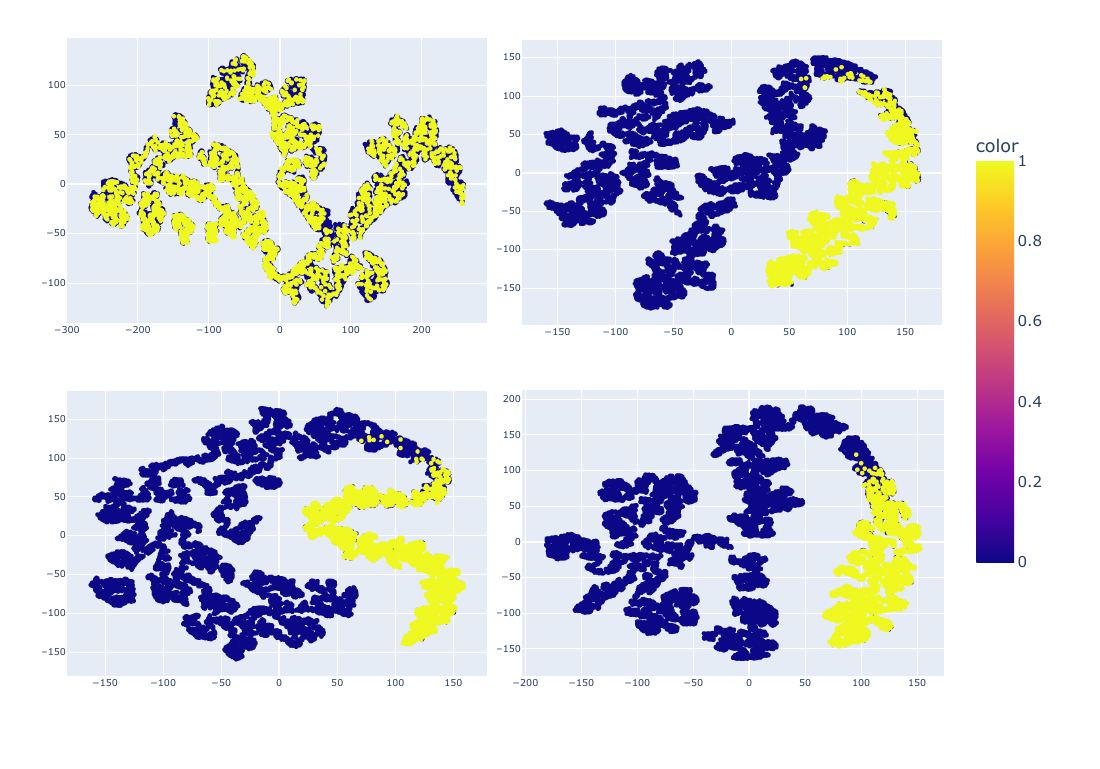}
\caption{This figure shows the feature distributions visual analysis of the UniAttackData~\cite{fang2024unified} with ResNet-50~\cite{lecun1998lenet} (left-top), vanilla CLIP~\cite{clip} (right-top), CLIP + Soft MoE (left-bottom), and CLIP + La-SoftMoE (right-bottom).}
\label{figure_ablation}
\hfil
\end{figure}

In order to reflect the effectiveness of our improvements more intuitively, we visually analyze the feature distributions of the UniAttackData~\cite{fang2024unified} with ResNet-50~\cite{lecun1998lenet}, vanilla CLIP~\cite{clip}, CLIP + Soft MoE, and CLIP + La-SoftMoE by t-SNE~\cite{van2008visualizing}.
As shown in Figure~\ref{figure_ablation}, the left-top demonstrates the feature distribution with ResNet-50, which shows the original distribution of the dataset is significant clustering and connectivity.
The right-top vanilla CLIP could classify images as fake or live faces, but the border between these two is not clear enough, some live ones are mixed in the fake cluster.
After combining the Soft MoE into the CLIP, the border is clearer but not regular as shown in the left-bottom, increasing the complexity of the models.
The right-bottom shows our proposed La-SoftMoE CLIP, which not only has the clearest border but also has the most regular shape. It proves the effectiveness of our method in the feature space.

 \begin{table}
     \centering
     \resizebox{\linewidth}{!}{
     \begin{tabular}{c|cccc}
         \hline
          Method	&ACER(\%)$\downarrow$	&ACC(\%)$\uparrow$	&AUC(\%)$\uparrow$	&EER(\%)$\downarrow$ \\
         \hline
          vanilla CLIP~\cite{clip}	        &0.91-	&98.87-	&\textbf{99.76}-	&0.96-\\
          CLIP + Soft MoE	&0.53$\downarrow$	&99.39$\uparrow$	&99.66$\downarrow$	&0.68$\downarrow$\\ 
          CLIP + La-SoftMoE &\textbf{0.32}$\downarrow$	&\textbf{99.54}$\uparrow$	&99.72$\uparrow$	&\textbf{0.56}$\downarrow$\\
         \hline
     \end{tabular}}
     \caption{
     The ablation results in vanilla CLIP~\cite{clip}, CLIP + Soft MoE, and CLIP + La-SoftMoE with the UniAttackData~\cite{fang2024unified}.
     }
     \label{result_abl}
 \end{table}

\section{Conclusion and Future Work}

In this paper, we proposed a novel La-SoftMoE CLIP, which allows for more flexible adaptation to the UAD task, significantly enhancing the model's capability to handle diversity attacks.
To prove the ability and effectiveness of our La-SoftMoE, we analyze it on both UniAttackData and JFSFD datasets and show its reliability in UAD tasks. 
In future work, we would consider to further improve the generalization of our method and explore the impact of prompt sentences on performance, which influences our model a lot.
And we would try to find a new weighting mechanism to make the model's self-adapting better.

\section*{Acknowledgment}
This work was supported by the National Key Research and Development Plan under Grant 2021YFF0602103, the China Postdoctoral Science Foundation 2023M743756, Beijing Natural Science Foundation JQ23016, the Science and Technology Development Fund of Macau Project 0123/2022/A3, 0070/2020/AMJ, 0096/2023/RIA2, and CCF-Zhipu AI Large Model OF 202219, the Chinese National Natural Science Foundation Projects U23B2054, 62276254, and InnoHK program.
{\small
\bibliographystyle{ieee}
\bibliography{egbib}

\begin{thebibliography}{10}\itemsep=-1pt

\bibitem{chen2020simpclr}
T.~Chen, S.~Kornblith, M.~Norouzi, and G.~Hinton.
\newblock A simple framework for contrastive learning of visual representations.
\newblock In {\em International conference on machine learning}, pages 1597--1607. PMLR, 2020.

\bibitem{ciftci2020fakecatcher}
U.~A. Ciftci, I.~Demir, and L.~Yin.
\newblock Fakecatcher: Detection of synthetic portrait videos using biological signals.
\newblock {\em IEEE transactions on pattern analysis and machine intelligence}, 2020.

\bibitem{dang2020detection}
H.~Dang, F.~Liu, J.~Stehouwer, X.~Liu, and A.~K. Jain.
\newblock On the detection of digital face manipulation.
\newblock In {\em Proceedings of the IEEE/CVF Conference on Computer Vision and Pattern recognition}, pages 5781--5790, 2020.

\bibitem{deb2023unified}
D.~Deb, X.~Liu, and A.~K. Jain.
\newblock Unified detection of digital and physical face attacks.
\newblock In {\em 2023 IEEE 17th International Conference on Automatic Face and Gesture Recognition (FG)}, pages 1--8. IEEE, 2023.

\bibitem{dolhansky2019deepfake}
B.~Dolhansky, R.~Howes, B.~Pflaum, N.~Baram, and C.~C. Ferrer.
\newblock The deepfake detection challenge (dfdc) preview dataset.
\newblock {\em arXiv preprint arXiv:1910.08854}, 2019.

\bibitem{ViT}
A.~Dosovitskiy, L.~Beyer, A.~Kolesnikov, D.~Weissenborn, X.~Zhai, T.~Unterthiner, M.~Dehghani, M.~Minderer, G.~Heigold, S.~Gelly, et~al.
\newblock An image is worth 16x16 words: Transformers for image recognition at scale.
\newblock {\em arXiv preprint arXiv:2010.11929}, 2020.

\bibitem{erdogmus20143DMAD}
N.~Erdogmus and S.~Marcel.
\newblock Spoofing face recognition with 3d masks.
\newblock {\em IEEE transactions on information forensics and security}, 9(7):1084--1097, 2014.

\bibitem{Fang_2023_CVPR}
H.~Fang, A.~Liu, J.~Wan, S.~Escalera, H.~J. Escalante, and Z.~Lei.
\newblock Surveillance face presentation attack detection challenge.
\newblock In {\em Proceedings of the IEEE/CVF Conference on Computer Vision and Pattern Recognition (CVPR) Workshops}, pages 6360--6370, June 2023.

\bibitem{fang2023surveillance}
H.~Fang, A.~Liu, J.~Wan, S.~Escalera, C.~Zhao, X.~Zhang, S.~Z. Li, and Z.~Lei.
\newblock Surveillance face anti-spoofing.
\newblock {\em IEEE Transactions on Information Forensics and Security}, 2023.

\bibitem{fang2024unified}
H.~Fang, A.~Liu, H.~Yuan, J.~Zheng, D.~Zeng, Y.~Liu, J.~Deng, S.~Escalera, X.~Liu, J.~Wan, and Z.~Lei.
\newblock Unified physical-digital face attack detection, 2024.

\bibitem{fedus2022review}
W.~Fedus, J.~Dean, and B.~Zoph.
\newblock A review of sparse expert models in deep learning.
\newblock {\em arXiv preprint arXiv:2209.01667}, 2022.

\bibitem{guo2023semantic}
G.~Guo, L.~Han, L.~Wang, D.~Zhang, and J.~Han.
\newblock Semantic-aware knowledge distillation with parameter-free feature uniformization.
\newblock {\em Visual Intelligence}, 1(1):6, 2023.

\bibitem{he2016deep}
K.~He, X.~Zhang, S.~Ren, and J.~Sun.
\newblock Deep residual learning for image recognition.
\newblock In {\em Proceedings of the IEEE conference on computer vision and pattern recognition}, pages 770--778, 2016.

\bibitem{jacobs1991adaptive}
R.~A. Jacobs, M.~I. Jordan, S.~J. Nowlan, and G.~E. Hinton.
\newblock Adaptive mixtures of local experts.
\newblock {\em Neural computation}, 3(1):79--87, 1991.

\bibitem{jiang2024effectiveness}
Y.~Jiang, X.~Yan, G.-P. Ji, K.~Fu, M.~Sun, H.~Xiong, D.-P. Fan, and F.~S. Khan.
\newblock Effectiveness assessment of recent large vision-language models.
\newblock {\em Visual Intelligence}, 2(17), 2024.

\bibitem{linearattention}
A.~Katharopoulos, A.~Vyas, N.~Pappas, and F.~Fleuret.
\newblock Transformers are rnns: Fast autoregressive transformers with linear attention.
\newblock In {\em Proceedings of the International Conference on Machine Learning (ICML)}, 2020.

\bibitem{korshunov2018deepfakes}
P.~Korshunov and S.~Marcel.
\newblock Deepfakes: a new threat to face recognition? assessment and detection.
\newblock {\em arXiv preprint arXiv:1812.08685}, 2018.

\bibitem{lecun1998lenet}
Y.~LeCun, L.~Bottou, Y.~Bengio, and P.~Haffner.
\newblock Gradient-based learning applied to document recognition.
\newblock {\em Proceedings of the IEEE}, 86(11):2278--2324, 1998.

\bibitem{lepikhin2020gshard}
D.~Lepikhin, H.~Lee, Y.~Xu, D.~Chen, O.~Firat, Y.~Huang, M.~Krikun, N.~Shazeer, and Z.~Chen.
\newblock Gshard: Scaling giant models with conditional computation and automatic sharding.
\newblock {\em arXiv preprint arXiv:2006.16668}, 2020.

\bibitem{li2023mimic}
B.~Li, Y.~Zhang, L.~Chen, J.~Wang, F.~Pu, J.~Yang, C.~Li, and Z.~Liu.
\newblock Mimic-it: Multi-modal in-context instruction tuning.
\newblock {\em arXiv preprint arXiv:2306.05425}, 2023.

\bibitem{li2018unsupervised}
H.~Li, W.~Li, H.~Cao, S.~Wang, F.~Huang, and A.~C. Kot.
\newblock Unsupervised domain adaptation for face anti-spoofing.
\newblock {\em IEEE Transactions on Information Forensics and Security}, 13(7):1794--1809, 2018.

\bibitem{li2020celeb}
Y.~Li, X.~Yang, P.~Sun, H.~Qi, and S.~Lyu.
\newblock Celeb-df: A large-scale challenging dataset for deepfake forensics.
\newblock In {\em Proceedings of the IEEE/CVF conference on computer vision and pattern recognition}, pages 3207--3216, 2020.

\bibitem{liu2024moeit}
A.~Liu.
\newblock Ca-moeit: Generalizable face anti-spoofing via dual cross-attention and semi-fixed mixture-of-expert.
\newblock {\em International Journal of Computer Vision}, pages 1--14, 2024.

\bibitem{liu2021cross}
A.~Liu, X.~Li, J.~Wan, Y.~Liang, S.~Escalera, H.~J. Escalante, M.~Madadi, Y.~Jin, Z.~Wu, X.~Yu, et~al.
\newblock Cross-ethnicity face anti-spoofing recognition challenge: A review.
\newblock {\em IET Biometrics}, 10(1):24--43, 2021.

\bibitem{ijcai2022p165}
A.~Liu and Y.~Liang.
\newblock Ma-vit: Modality-agnostic vision transformers for face anti-spoofing.
\newblock In {\em Proceedings of the Thirty-First International Joint Conference on Artificial Intelligence, {IJCAI-22}}, pages 1180--1186, 2022.

\bibitem{CeFAdatabase2021}
A.~Liu, Z.~Tan, J.~Wan, S.~Escalera, G.~Guo, and S.~Z. Li.
\newblock Casia-surf cefa: A benchmark for multi-modal cross-ethnicity face anti-spoofing.
\newblock In {\em 2021 IEEE Winter Conference on Applications of Computer Vision (WACV)}, Jan 2021.

\bibitem{liu2021face}
A.~Liu, Z.~Tan, J.~Wan, Y.~Liang, Z.~Lei, G.~Guo, and S.~Z. Li.
\newblock Face anti-spoofing via adversarial cross-modality translation.
\newblock {\em IEEE Transactions on Information Forensics and Security}, 16:2759--2772, 2021.

\bibitem{liu2023fm}
A.~Liu, Z.~Tan, Z.~Yu, C.~Zhao, J.~Wan, Y.~L.~Z. Lei, D.~Zhang, S.~Z. Li, and G.~Guo.
\newblock Fm-vit: Flexible modal vision transformers for face anti-spoofing.
\newblock {\em IEEE Transactions on Information Forensics and Security}, 2023.

\bibitem{liu2019multi}
A.~Liu, J.~Wan, S.~Escalera, H.~Jair~Escalante, Z.~Tan, Q.~Yuan, K.~Wang, C.~Lin, G.~Guo, I.~Guyon, et~al.
\newblock Multi-modal face anti-spoofing attack detection challenge at cvpr2019.
\newblock In {\em Proceedings of the IEEE/CVF conference on computer vision and pattern recognition workshops}, pages 0--0, 2019.

\bibitem{liu2024cfpl}
A.~Liu, S.~Xue, J.~Gan, J.~Wan, Y.~Liang, J.~Deng, S.~Escalera, and Z.~Lei.
\newblock Cfpl-fas: Class free prompt learning for generalizable face anti-spoofing.
\newblock In {\em Proceedings of the IEEE/CVF Conference on Computer Vision and Pattern Recognition}, pages 222--232, 2024.

\bibitem{liu20213d}
A.~Liu, C.~Zhao, Z.~Yu, A.~Su, X.~Liu, Z.~Kong, J.~Wan, S.~Escalera, H.~J. Escalante, Z.~Lei, et~al.
\newblock 3d high-fidelity mask face presentation attack detection challenge.
\newblock In {\em Proceedings of the IEEE/CVF International Conference on Computer Vision Workshops}, pages 814--823, 2021.

\bibitem{liu2022contrastive}
A.~Liu, C.~Zhao, Z.~Yu, J.~Wan, A.~Su, X.~Liu, Z.~Tan, S.~Escalera, J.~Xing, Y.~Liang, et~al.
\newblock Contrastive context-aware learning for 3d high-fidelity mask face presentation attack detection.
\newblock {\em IEEE Transactions on Information Forensics and Security}, 17:2497--2507, 2022.

\bibitem{liu20163d}
S.~Liu, P.~C. Yuen, S.~Zhang, and G.~Zhao.
\newblock 3d mask face anti-spoofing with remote photoplethysmography.
\newblock In {\em Computer Vision--ECCV 2016: 14th European Conference, Amsterdam, The Netherlands, October 11--14, 2016, Proceedings, Part VII 14}, pages 85--100. Springer, 2016.

\bibitem{liu2018SiW}
Y.~Liu, A.~Jourabloo, and X.~Liu.
\newblock Learning deep models for face anti-spoofing: Binary or auxiliary supervision.
\newblock In {\em Proceedings of the IEEE conference on computer vision and pattern recognition}, pages 389--398, 2018.

\bibitem{liu2018learning}
Y.~Liu, A.~Jourabloo, and X.~Liu.
\newblock Learning deep models for face anti-spoofing: Binary or auxiliary supervision.
\newblock In {\em Proceedings of the IEEE conference on computer vision and pattern recognition}, pages 389--398, 2018.

\bibitem{puigcerver2023sparse}
J.~Puigcerver, C.~Riquelme, B.~Mustafa, and N.~Houlsby.
\newblock From sparse to soft mixtures of experts.
\newblock {\em arXiv preprint arXiv:2308.00951}, 2023.

\bibitem{clip}
A.~Radford, J.~W. Kim, C.~Hallacy, A.~Ramesh, G.~Goh, S.~Agarwal, G.~Sastry, A.~Askell, P.~Mishkin, J.~Clark, et~al.
\newblock Learning transferable visual models from natural language supervision.
\newblock In {\em International conference on machine learning}, pages 8748--8763. PMLR, 2021.

\bibitem{rossler2019faceforensics++}
A.~Rossler, D.~Cozzolino, L.~Verdoliva, C.~Riess, J.~Thies, and M.~Nie{\ss}ner.
\newblock Faceforensics++: Learning to detect manipulated facial images.
\newblock In {\em Proceedings of the IEEE/CVF international conference on computer vision}, pages 1--11, 2019.

\bibitem{sarkar2022gan}
E.~Sarkar, P.~Korshunov, L.~Colbois, and S.~Marcel.
\newblock Are gan-based morphs threatening face recognition?
\newblock In {\em ICASSP 2022-2022 IEEE International Conference on Acoustics, Speech and Signal Processing (ICASSP)}, pages 2959--2963. IEEE, 2022.

\bibitem{shazeer2017outrageously}
N.~Shazeer, A.~Mirhoseini, K.~Maziarz, A.~Davis, Q.~Le, G.~Hinton, and J.~Dean.
\newblock Outrageously large neural networks: The sparsely-gated mixture-of-experts layer.
\newblock {\em arXiv preprint arXiv:1701.06538}, 2017.

\bibitem{shen2021efficient}
Z.~Shen, M.~Zhang, H.~Zhao, S.~Yi, and H.~Li.
\newblock Efficient attention: Attention with linear complexities.
\newblock In {\em Proceedings of the IEEE/CVF winter conference on applications of computer vision}, pages 3531--3539, 2021.

\bibitem{srivatsan2023flip}
K.~Srivatsan, M.~Naseer, and K.~Nandakumar.
\newblock Flip: Cross-domain face anti-spoofing with language guidance.
\newblock In {\em Proceedings of the IEEE/CVF International Conference on Computer Vision}, pages 19685--19696, 2023.

\bibitem{tang2024wfss}
L.~Tang, Z.~Yin, H.~Su, W.~Lyu, and B.~Luo.
\newblock Wfss: weighted fusion of spectral transformer and spatial self-attention for robust hyperspectral image classification against adversarial attacks.
\newblock {\em Visual Intelligence}, 2(1):5, 2024.

\bibitem{van2008visualizing}
L.~Van~der Maaten and G.~Hinton.
\newblock Visualizing data using t-sne.
\newblock {\em Journal of machine learning research}, 9(11), 2008.

\bibitem{vaswani2017attention}
A.~Vaswani, N.~Shazeer, N.~Parmar, J.~Uszkoreit, L.~Jones, A.~N. Gomez, {\L}.~Kaiser, and I.~Polosukhin.
\newblock Attention is all you need.
\newblock {\em Advances in neural information processing systems}, 30, 2017.

\bibitem{wang2020linformer}
S.~Wang, B.~Z. Li, M.~Khabsa, H.~Fang, and H.~Ma.
\newblock Linformer: Self-attention with linear complexity.
\newblock {\em arXiv preprint arXiv:2006.04768}, 2020.

\bibitem{wen2015face}
D.~Wen, H.~Han, and A.~K. Jain.
\newblock Face spoof detection with image distortion analysis.
\newblock {\em IEEE Transactions on Information Forensics and Security}, 10(4):746--761, 2015.

\bibitem{yang2023continual}
Y.~Yang, Z.~Cui, J.~Xu, C.~Zhong, W.-S. Zheng, and R.~Wang.
\newblock Continual learning with bayesian model based on a fixed pre-trained feature extractor.
\newblock {\em Visual Intelligence}, 1(1):5, 2023.

\bibitem{yin2023survey}
S.~Yin, C.~Fu, S.~Zhao, K.~Li, X.~Sun, T.~Xu, and E.~Chen.
\newblock A survey on multimodal large language models.
\newblock {\em arXiv preprint arXiv:2306.13549}, 2023.

\bibitem{yu2023visual}
Z.~Yu, R.~Cai, Y.~Cui, A.~Liu, and C.~Chen.
\newblock Visual prompt flexible-modal face anti-spoofing.
\newblock {\em arXiv preprint arXiv:2307.13958}, 2023.

\bibitem{yu2024benchmarking}
Z.~Yu, R.~Cai, Z.~Li, W.~Yang, J.~Shi, and A.~C. Kot.
\newblock Benchmarking joint face spoofing and forgery detection with visual and physiological cues.
\newblock {\em IEEE Transactions on Dependable and Secure Computing}, 2024.

\bibitem{yu2020face}
Z.~Yu, X.~Li, X.~Niu, J.~Shi, and G.~Zhao.
\newblock Face anti-spoofing with human material perception.
\newblock In {\em Computer Vision--ECCV 2020: 16th European Conference, Glasgow, UK, August 23--28, 2020, Proceedings, Part VII 16}, pages 557--575. Springer, 2020.

\bibitem{yu2019remote}
Z.~Yu, W.~Peng, X.~Li, X.~Hong, and G.~Zhao.
\newblock Remote heart rate measurement from highly compressed facial videos: an end-to-end deep learning solution with video enhancement.
\newblock In {\em Proceedings of the IEEE/CVF international conference on computer vision}, pages 151--160, 2019.

\bibitem{yu2022deep}
Z.~Yu, Y.~Qin, X.~Li, C.~Zhao, Z.~Lei, and G.~Zhao.
\newblock Deep learning for face anti-spoofing: A survey.
\newblock {\em IEEE transactions on pattern analysis and machine intelligence}, 45(5):5609--5631, 2022.

\bibitem{yu2020fas}
Z.~Yu, J.~Wan, Y.~Qin, X.~Li, S.~Z. Li, and G.~Zhao.
\newblock Nas-fas: Static-dynamic central difference network search for face anti-spoofing.
\newblock {\em IEEE transactions on pattern analysis and machine intelligence}, 43(9):3005--3023, 2020.

\bibitem{yu2020searching}
Z.~Yu, C.~Zhao, Z.~Wang, Y.~Qin, Z.~Su, X.~Li, F.~Zhou, and G.~Zhao.
\newblock Searching central difference convolutional networks for face anti-spoofing.
\newblock In {\em Proceedings of the IEEE/CVF conference on computer vision and pattern recognition}, pages 5295--5305, 2020.

\bibitem{yuan2024unified}
H.~Yuan, A.~Liu, J.~Zheng, J.~Wan, J.~Deng, S.~Escalera, H.~J. Escalante, I.~Guyon, and Z.~Lei.
\newblock Unified physical-digital attack detection challenge, 2024.

\bibitem{zeng2024human}
Q.~Zeng, Y.~Xie, Z.~Lu, and Y.~Xia.
\newblock A human-in-the-loop method for pulmonary nodule detection in ct scans.
\newblock {\em Visual Intelligence}, 2(1):1--13, 2024.

\bibitem{zhang2020casia}
S.~Zhang, A.~Liu, J.~Wan, Y.~Liang, G.~Guo, S.~Escalera, H.~J. Escalante, and S.~Z. Li.
\newblock Casia-surf: A large-scale multi-modal benchmark for face anti-spoofing.
\newblock {\em TBMIO}, 2(2):182--193, 2020.

\bibitem{zhang2023multimodal}
Z.~Zhang, A.~Zhang, M.~Li, H.~Zhao, G.~Karypis, and A.~Smola.
\newblock Multimodal chain-of-thought reasoning in language models.
\newblock {\em arXiv preprint arXiv:2302.00923}, 2023.

\bibitem{zhao2021multi}
H.~Zhao, W.~Zhou, D.~Chen, T.~Wei, W.~Zhang, and N.~Yu.
\newblock Multi-attentional deepfake detection.
\newblock In {\em Proceedings of the IEEE/CVF conference on computer vision and pattern recognition}, pages 2185--2194, 2021.

\bibitem{zhou2022conditional}
K.~Zhou, J.~Yang, C.~C. Loy, and Z.~Liu.
\newblock Conditional prompt learning for vision-language models.
\newblock In {\em Proceedings of the IEEE/CVF conference on computer vision and pattern recognition}, pages 16816--16825, 2022.

\bibitem{zhou2022learning}
K.~Zhou, J.~Yang, C.~C. Loy, and Z.~Liu.
\newblock Learning to prompt for vision-language models.
\newblock {\em International Journal of Computer Vision}, 130(9):2337--2348, 2022.

\bibitem{zhou2017two}
P.~Zhou, X.~Han, V.~I. Morariu, and L.~S. Davis.
\newblock Two-stream neural networks for tampered face detection.
\newblock In {\em 2017 IEEE conference on computer vision and pattern recognition workshops (CVPRW)}, pages 1831--1839. IEEE, 2017.

\bibitem{zoph2022st}
B.~Zoph, I.~Bello, S.~Kumar, N.~Du, Y.~Huang, J.~Dean, N.~Shazeer, and W.~Fedus.
\newblock St-moe: Designing stable and transferable sparse expert models.
\newblock {\em arXiv preprint arXiv:2202.08906}, 2022.

\end{thebibliography}
}

\end{document}